\documentclass[letterpaper,twocolumn,10pt]{article}
\usepackage{usenix,epsfig,endnotes}

\usepackage{times}
\usepackage{algorithmic}
\usepackage{algorithm}
\usepackage{epsfig}
\usepackage{amssymb}
\usepackage{amsmath}
\usepackage{amsfonts}
\usepackage{subfigure}
\usepackage{nccmath}
\usepackage{graphicx}
\usepackage{url}
\usepackage{multirow}
\usepackage{wrapfig}
\usepackage{caption}
\usepackage{subfigure}
\usepackage{caption}
\usepackage{color}
\usepackage{enumitem}

\newcommand{\dist}{\mbox{dist}}
\newcommand{\seuc}{\mbox{seuc}}
\newcommand{\tripd}{\mbox{tripd}}
\newcommand{\simf}{\mbox{sim}}

\begin{document}

\title{\Large \bf Trip Prediction by Leveraging Trip Histories from Neighboring Users}

\author{
{\rm Yuxin Chen}\\
\normalsize Department of Computer Science \\ 
\normalsize  University of Chicago, USA \\
\rm \normalsize  \texttt{chenyuxin@uchicago.edu}
\and
{\rm Morteza Haghir Chehreghani}\\
\normalsize Department of Computer Science and Engineering \\ 
\normalsize Chalmers University of Technology, Sweden \\
\rm \normalsize  \texttt{morteza.chehreghani@chalmers.se}
}

\maketitle

\begin{abstract}
  We propose a novel approach for trip prediction by analyzing user's trip histories. We augment users' (self-) trip histories by adding ``similar'' trips from other users, which could be informative and useful for predicting future trips for a given user. This also helps to cope with noisy or sparse trip histories, where the self-history by itself does not provide a reliable prediction of future trips. We show empirical evidence that by enriching the users' trip histories with additional trips, one can improve the prediction error by 15\%$\sim$40\%, evaluated on multiple subsets of the \texttt{Nancy2012} dataset. This real-world dataset is collected from public transportation ticket validations in the city of Nancy, France. Our prediction tool is a central component of a  trip simulator system designed to analyze the functionality of public transportation in the city of Nancy.
\end{abstract}


\section{Introduction and Background}

This paper is concerned with prediction of future trips according to trip histories. Trip prediction is used, for example, to simulate a public transportation system, to analyze the traffic, to investigate the demand and load, and to identify the bottlenecks and the constraints. In principle, a universal trip planner can be used to predict a trip for a user, 
by sampling from the predictive distribution of future trips, marginalized over the entire population. Such trips usually reflect the general traveling pattern of all the users in the system, and are useful for estimating the overall statistics of a transportation system. However, passengers do not always travel according to what a trip planner offers to them, due to the fact that they have different trip preferences and therefore behave differently according to various criteria. Thereby, looking at individual historical trips can provide very useful information about trip habits and behavior of them, and hence is helpful for making personalized recommendations.

\paragraph{Personalized trip prediction}
A partially related task to trip prediction is trip recommendation, where a recommendation system deals with the \emph{choice of the routes} to be taken by the user. A trip predictor, on the other hand, aims at \emph{predicting the trip} that user $u$ will take at time $t$ (i.e. to estimate the origin and the destination).
Training a trip prediction model at individual level is essential for building trip planner systems. This is due to the fact that personalized trip planner \cite{letchner2006trip,ricci2011introduction,Wang:2014:RRR:2733004.2733027,Chidlovskii:MUD22015,doi:10.1080/01441640500333677,arentze2013adaptive} recommends future trips for a given user, by sampling from the predictive distribution. The sampling is conditioning on the query and the historical behaviors of the user (i.e., $\text{target} \sim \Pr(\text{trip} \mid \text{query}, \text{history})$). The query (e.g., ``what's the most popular trip on Monday afternoon'', ``best route to work in two hours'') is universal for all users, whereas the history usually consists of user's profiles \cite{park2007location}, real-time trip information (e.g., traffic \cite{letchner2006trip} and real-time location \cite{Wang:2014:RRR:2733004.2733027}), community-contributed meta-data (e.g., geo-tags \cite{kurashima2010travel} and photos \cite{majid2013context}, etc), previously realized trips (e.g., trips realized in a different city \cite{clements2011personalised,majid2013context}) and so on, which encodes user's bias / preference towards specific trips. Such information is crucial in steering the trip recommender towards making suggestions that are meaningful to the user, as well as predicting future trips. For example, students who travel to schools often take trips with the shortest travel time; while parents who pick up their children at school may very likely choose a different route, so that they can drop by local shops and do grocery shopping on the way. If we observe a direct trip from location $A$ (e.g., home) to $B$ (e.g., school) frequently enough in the travel history of a given user, it is very likely for the same user to take this trip under similar conditions,
as we may guess she is a user of some fixed type (e.g., ``students''). With sufficient amount of historical data from each user, learning their attributes (i.e., latent variables) \cite{Cheng:2011:PTR:2072298.2072311,arentze2013adaptive} from trip histories is helpful for us to train a good trip predictor. In this way, it helps to group users of common interest, which in turn regularizes the model to make meaningful  predictions \cite{ekstrand2011collaborative}.

\begin{figure*}[htb]
  \centering
  \hspace{-5mm}
  \includegraphics[width=0.77\textwidth]{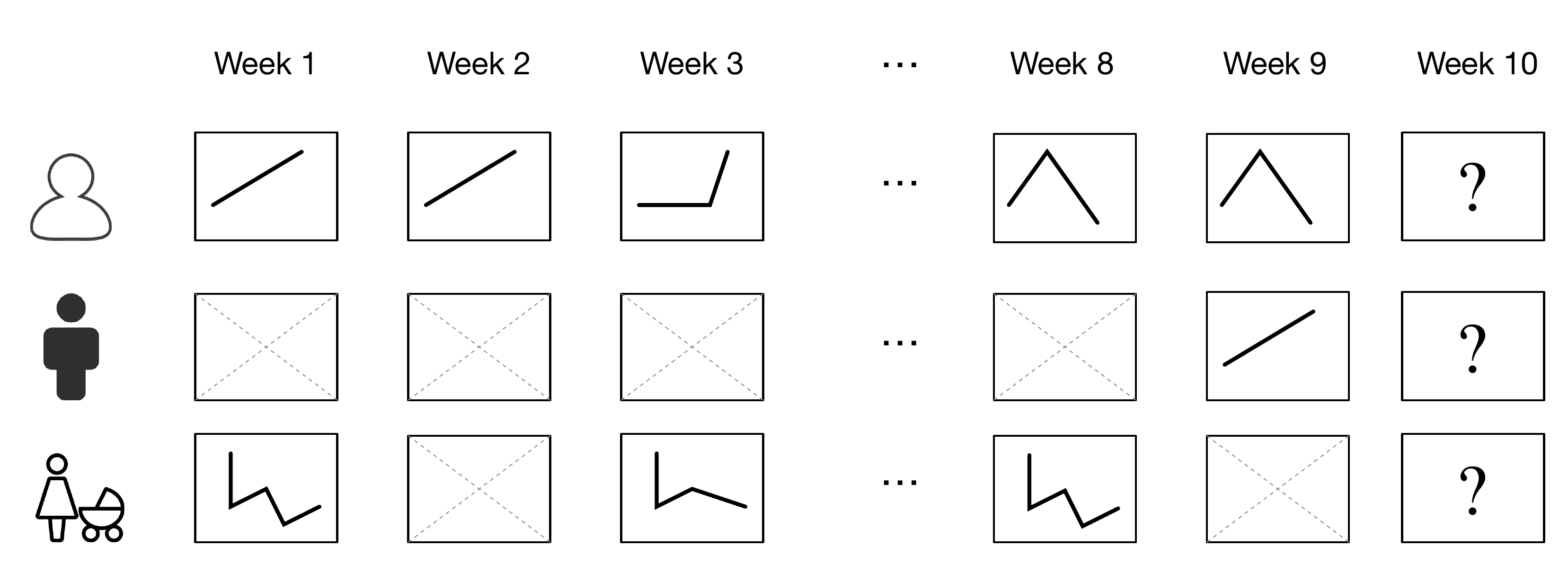}
  \caption{Illustration of trips taken by users at some fixed time slots. We can see the histories of trips taken by three different users, one for each line, on the same time slots (say, from 8am to 9am on Mondays) over 9 weeks. The line segments inside the rectangles represent trajectories of the trips; the turning points represent public transport transit stops. A trip predictor aims to predict the trips for each of the users at some given time slot, e.g., from 8am to 9am, on Monday of week 10.}
  \label{fig:trip_illustration}
\end{figure*}

One particular challenge for trip prediction is the ``cold-start'' problem: we may want to predict the behavior of very recent or newly-joined users, who do not yet have a long travel history in the system. It is challenging to represent such users (i.e., finding expressive representation in vector space) purely based on histories of their own, since there is not sufficient knowledge about them.
As an example, the user in second line of Figure~\ref{fig:trip_illustration} has a rather short trip history, and the single trip it takes on week 9 only contains limited information about the user's behavior.
 In such scenarios, one often needs to trade exploration (i.e., collecting more information about the user) and exploitation (i.e., making prediction from the realized trips).\footnote{One common setting to study this problem is online learning, where one seeks to iteratively predict trips for a given user and explore new knowledge, aiming to maximize the cumulative relevance of the entire set \cite{Akerblom0C20,abs-2111-02314,abs-2109-08467}.} However, in this work we focus on \emph{exploitation} of information which is already known and present, and we target at maximizing the relevance of the current predictions. We hope to develop an intelligent trip prediction system that is capable of decoding users' travel patterns from historical travel data, even when the trip history is not statistically rich enough.

\paragraph{Similarity-based approach}
A natural way to address such ``cold-start'' problem is to enrich users' trip history by looking into historical trips of ``similar'' users. Similarity-based approach has been popular as an alternative to traditional feature-based learning for numerous machine learning tasks, especially when it is challenging to find satisfactory (vectorial representation of) features for learning purposes \cite{wilson2000reduction}. A commonly used approach to similarity-based classification is $k$-nearest-neighbors (k-NN). Nearest-neighbor learning is the algorithmic analogue of the exemplar model of human learning \cite{goldstone2003concepts}. Although simple, it is often effective in practice, and empirical studies on a number of benchmark data sets show that it is difficult to surpass the performance of k-NN for similarity data \cite{pekalska2002generalized}.

We build upon and extend the nearest-neighbor paradigm to personalized trip prediction problem. One fundamental problem we face is to develop a proper similarity measurement between different users. Optimizing the similarity measure has been extensively studied in both supervised learning and unsupervised learning setting. Under the supervised learning setting, where target trips are known, one can optimize the similarity measure via metric learning \cite{yang2006distance}. In unsupervised learning setting, one can regularize the models via, e.g., non-negative matrix factorization \cite{lee2001algorithms,hastie2009unsupervised}. In both settings, however, the goal is to learn a similarity / distance measurement that leads to \emph{optimal clustering}, rather than finding the most useful users which incur the minimal prediction error.
In contrast, we take a direct approach, where we directly relate the similarity between two users with the performance w.r.t. the prediction error. We argue that our work is orthogonal to existing methods that learn feature representations, and we can always learn better features via these methods as pre-processing steps for our approach.

One work is relevant to~\cite{jiang2009improving}, where the goal is to find clusters of users to minimize the prediction error. In \cite{jiang2009improving} the authors make a direct connection between user groups and prediction error: a user is assigned to a group as long as it helps reducing prediction error, and the micro-segmentation of user groups are used \emph{collectively} for targeted and personalized prediction. Despite of that, we do not require partitioning the users into groups; instead, we allow the similarity measure to be asymmetric, where \emph{neighbors} (i.e., similar users) of a given user do not necessarily have that user as a neighbor. Such relaxation provides us with flexibility of creating arbitrary groups that could help with reducing the expected prediction error.

\paragraph{Our contribution} To sum up, our contributions include the following aspects:
\\
1. We develop a general framework for trip prediction, which predicts future trips for given users by incorporating their historical trip choices. Unlike existing works \cite{letchner2006trip,Wang:2014:RRR:2733004.2733027,kurashima2010travel,majid2013context} 
  which exploit only the user-specific information, we focus on developing a framework on top of such systems, which aims at improving the performance (i.e., prediction error) of such systems by leveraging historical trips of \emph{similar users}, where our similarity measurement by definition is related to the reduction in prediction error in a separate validation set.
  \\
2. We demonstrate the effectiveness of our method on multiple subsets of the \texttt{Nancy2012} public transportation dataset, and show that by enriching the users' trip histories with additional trips, one can improve the performance of trip predictor by 15\%$\sim$40\% in terms of prediction error. An important aspect of our study is that the dataset is real-world, i.e., it is collected from real (electronic) trip transactions of the passengers in the city of Nancy, France, during the year 2012. By predicting the trips at an individual level, our prediction tool constitutes an important component of a  trip simulator system, which is designed and implemented to analyze the functionality of public transportation and the behavior of users in the city of Nancy.
    For instance, the trip simulator might query the future trips of all users in the system at a specific time.\footnote{This information can be obtained at different time points which can help to for example understand the dynamics of traffic and the functionality of transportation system \cite{cats2011dynamic,moreira2016time}.} This leads to computing sufficient statistics of the overall traffic of the city, as well as zone specific traffic, by aggregating individual predicted trips. Such information can be used, for example, to detect the bottleneck of the system, decide whether bus stops should be added / placed, and support strategic decision making (e.g., to build new transportation system).
    \\
3. We discuss the possibility that further improvements in performance can be attained by (1) leveraging new, informative and fine-grade features and (2) transferring existing features into more robust representations via e.g. non-negative matrix factorization.

\section{Neighbor-based Trip Prediction}

Our goal is to develop an effective approach for trip prediction based on trip histories. Formally, we define a trip to be a tuple $(o,d,v)$, where $o$ denotes the (location of the) origin, $d$ denotes the destination, and $v$ denotes the list of transit stops (or via-points). Similarly, we use $(o,d,v)_{u,t}$ to denote a trip taken by user $u$ at $t$, where $t$ denotes the time of the trip\footnote{For public transportation, we use $t$ to denote the time in the week, e.g., 8am-9am on Monday, since the trips are often repetitive over different weeks. We use $o, d$ to denote locations of the origin and destination stops; this allows us to compute distances between two trips efficiently, and hence we don’t need to further discretize the space (e.g., using close-by bus stops or city areas) as is done in \cite{moreira2016time} (such quasi-discretization may lead to less accurate prediction in addition to extra computation time).
}. Individual users' trip histories might be sparse or noisy, thus, they might not be sufficient to provide a suitable feature representation for predicting future trips. Therefore, we need to augment the individual users' self trip histories by trip histories of other users  in order to compute a more robust estimation. However, taking all other user trip histories into account, i.e. averaging over all trips, is not appropriate, since different people might have different trip preferences and thus global averaging discards such a diversity. Therefore, for each user, we need to identify an additional set of trips histories (i.e. neighbors) than the self history which help to improve future trip prediction. To do so, we take two important considerations into account:
\\
I) The users usually make a diverse set of trips during a day. Therefore, it makes sense to divide a day into small (e.g., one-hour) time intervals and consider the trips inside this interval as unit of trip behavior. On the other hand, the trip behavior of user $u$ at time $t$ might be similar to the trip behavior of  user $u'$ at a different time $t'$ such that $t$ and $t'$ does not necessary overlap. For example, user $u$ might travel to the city university at time 9am, whereas user $u'$ might take this trip at time 3:00 pm. Therefore, when querying a trip as well as finding appropriate auxiliary trip histories, we parametrize the operations by time point $t$. Formally, we define the base \emph{entities} that we work with as user-time pairs, i.e., $e := \langle u,t\rangle$. We use $T_{ut}$ to refer to the set of trips associated with entity  $\langle u,t\rangle$:
  \begin{align*}
    T_{ut} = \left\{ (o,d,v)_{u',t'} : u'=u \wedge t'=t  \right\}
  \end{align*}
  Then,  the question is: \emph{for a specific entity $\langle u,t\rangle$ which represents user $u$ at time $t$, what are the other entities that can be used to obtain a better prediction for the next trip of the user, at time $t$ in future days/weeks?}
\\
II) The usefulness relations are not symmetric, i.e. entity $\langle u',t'\rangle$ may be helpful for entity $\langle u,t\rangle$ to find a better trip in future, however, $\langle u',t'\rangle$ might not need $\langle u,t\rangle$ for this purpose. In particular, such a directional relation can hold whenever the trip history of $\langle u',t'\rangle$ is clean and long enough, but $\langle u,t\rangle$ is very short or noisy. Thus, the methods that work based on grouping or clustering of entities discard this kind of asymmetric relations. 

Thereby, we propose a method to compute additional helpful entities to each specific entity. However, we do not have access to the user profile, such as meta-data description of the users' occupation, hobbies, and age group, etc. Instead, we are only given the users' trip histories, based on which we should define a proper time-dependent distance/similarity measure between them.
In absence of meta-data feature representation, we follow a similarity-based approach, while relying on the fundamental principle of learning theory: \emph{a good model should perform well on unseen data from the same source.}

\begin{figure}
  \centering
  \includegraphics[width=0.5\textwidth]{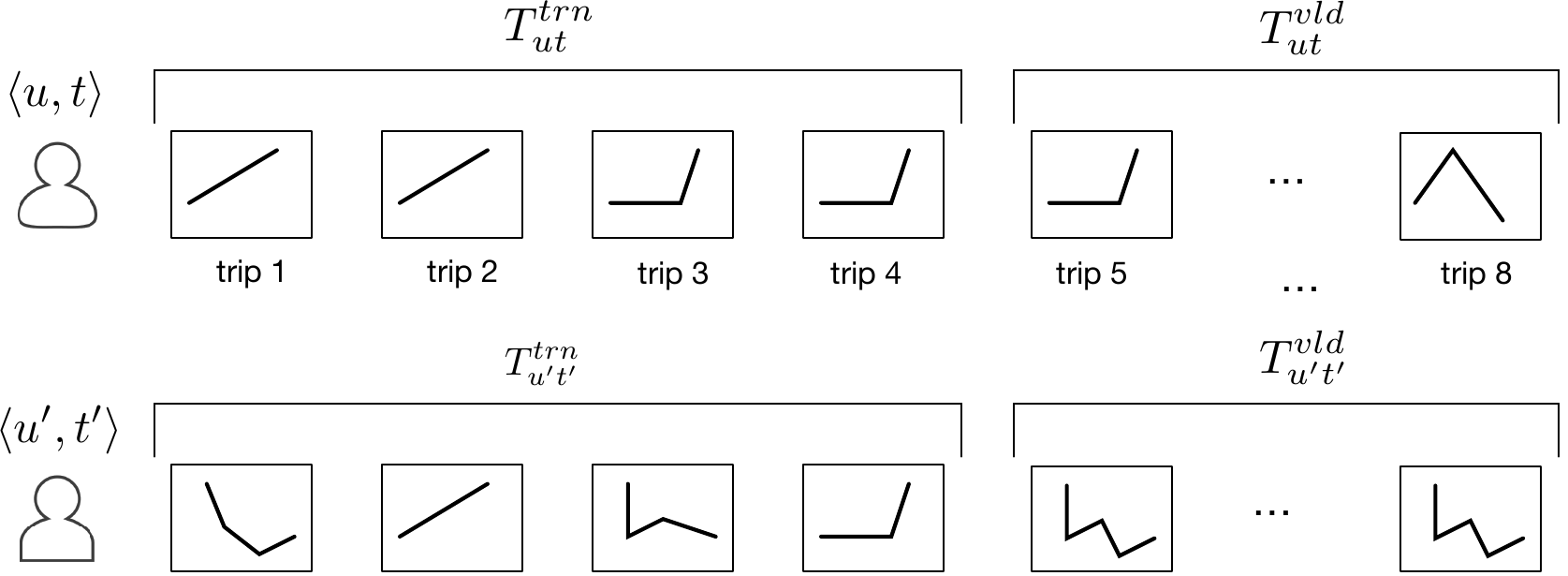}
  \caption{Computing neighboring entities of $\langle u,t\rangle$: we divide the dataset into training and validation set, each containing a fraction of trips for each entity.}
  \label{fig:tripsim_illustration}
\end{figure}

For a given entity $\langle u,t\rangle$, we determine its neighbors in a non-parametric way using a separate unseen dataset called \emph{validation} set. More formally, let us start with the simple case, where we are given a dataset of trip entities with equal trip histories $L$, i.e., $\mathbf D = \{\langle u,t\rangle: |T_{u,t}| = L\}$. We then divide the whole dataset into two subsets, training set and validation set, each containing $L/2$ trips per entity. Then, we use the validation set to identify the appropriate neighbors of the entities. For each $\langle u,t\rangle$ in the training set (resp. validation set), we denote the associated set of trips by $T_{ut}^{trn}$ (resp. $T_{ut}^{vld}$). See Figure~\ref{fig:tripsim_illustration} for illustration. In case of not availability of enough annotated data, active learning methods can be employed to provide annotated data in  a cost-efficient way \cite{Comuni2022,JARL2022104972}.

\begin{algorithm}[ht]
  \caption{History-based trip prediction.}
  \label{alg:trip_recom}
  \begin{algorithmic} [1]
    \REQUIRE {The entities and the respective trips.}
    \ENSURE Predicted trip(s) to each entity.
    \FOR{\textbf{each} entity $\langle u,t\rangle$}
    \STATE Split the trip histories into $T_{ut}^{trn}$ and $T_{ut}^{vld}$ for construction of the training and validation sets.
    \ENDFOR
    \FOR{\textbf{each} entity $\langle u,t\rangle$}
    \STATE $\mathcal N_{ut} = \{  \langle u',t'\rangle : \dist(T_{u't'}^{trn}, T_{ut}^{vld}) \le \dist(T_{ut}^{trn}, T_{ut}^{vld}) \}\, .$
    \STATE $r_{ut} \in \arg\max_{x\in T(\mathcal N_{ut})} \sum_{y \in T(\mathcal N_{ut})} f_x\; sim(x,y)\, .$
    \ENDFOR

    \STATE \textbf{return} $\{r_{ut}\}$
  \end{algorithmic}
\end{algorithm}

To compute the appropriate neighbors of the entity $\langle u,t\rangle$, we investigate which of the training histories are at least equally similar to the validation history  compared with the self training history, i.e.,
\begin{equation}
  \mathcal N_{ut} = \left\{  \langle u',t'\rangle : \dist(T_{u't'}^{trn}, T_{ut}^{vld}) \le \dist(T_{ut}^{trn}, T_{ut}^{vld}) \right\}
\end{equation}

where $\dist(.,.)$ denotes the distance between trip histories. In our study, we investigate two options for $\dist(.,.)$:

\begin{enumerate}[leftmargin=3mm]
\item \emph{ordered}, where only the trips at the same positions are compared, i.e.,
  \begin{equation}\label{eq:dist_ordered}
    \dist(p,q) = \frac{2}{L}\sum_{1 \le i \le L/2} \seuc(p_i,q_i)\, ,
  \end{equation}
  Hereby, we sort the trips in the trip histories according to their time of realization, and $p_i$ (resp. $q_i$) indicates the $i^{th}$ trip in trip history $p$ (resp. $q$). Further, $\seuc(p_i,q_i)$ gives the squared Euclidean distance between trips $p_i$ and $q_i$. Specifically, for two single-leg trips $p_i := (o_1,d_1,v)$ and $q_i := (o_2,d_2,v)$ where $v=\emptyset$, we have
  $\seuc(p_i, q_i) = \seuc\left(\langle o_1,d_1,v\rangle, \langle o_2,d_2,v\rangle \right) = |o_1-o_2|^2 + |d_1-d_2|^2.$
  Note that this variant requires $p$ and $q$ to include the same number of trips.

\item \emph{all2all}: where each trip from one trip history is compared against all trips of the other history, i.e.,
  \begin{equation}\label{eq:dist_all2all}
    \dist(p,q) = \frac{4}{L^2} \sum_{1 \le i \le L/2} \;\; \sum_{1 \le j \le L/2} \seuc(p_i,q_j)\, .
  \end{equation}
  One advantage of  \emph{all2all} over the \emph{ordered} variant is that $p$ and $q$ do not necessarily need to have the same number of trips. Thus, \emph{all2all} is more general-purpose and can be applied to trip histories with different length.
\end{enumerate}

We note that our framework is general enough to use other distances measures than Euclidean distances as well, e.g. the Minimax distances \cite{Chehreghani16SDM,ChehreghaniAAAI17}.
In the next step, we employ the members of the neighbor set $\mathcal N_{ut}$ to predict a future trip for entity $\langle u,t\rangle$. For this purpose, we consider the total trip histories of all neighbors collected in $\mathcal N_{ut}$ (i.e. including both training and validations trips) and compute the representative trip(s) as the trip(s) $r_{ut}$ with the minimal average distance (or maximal average similarity) with other trips. More precisely,

\begin{equation}
  r_{ut} \in \arg\max_{x\in T(\mathcal N_{ut})} \sum_{y \in T(\mathcal N_{ut})} f_x\cdot \simf(x,y)\, ,
\end{equation}

where $T(\mathcal N_{ut})$ indicates the set of all trips of all entities in $\mathcal N_{ut}$, and $f_x$ denotes the frequency of trip $x$ in this set. $\simf(x,y)$ measures the pairwise similarity between the two trips $x$ and $y$, which is obtained by $\texttt{const} - \seuc(x,y)$. We select $\texttt{const}$ as the minimal value for which the pairwise similarities become nonnegative. Finally, $r_{ut}$ is predicted as the next trip of the user under investigation. Note that $r_{ut}$ might not be deterministic (if there are ties among multiple trips). Algorithm~\ref{alg:trip_recom} describes the whole procedure in detail.


\begin{figure}[th!]
  \centering
  \subfigure[L=2]
  {
    \includegraphics[width=0.45\textwidth]{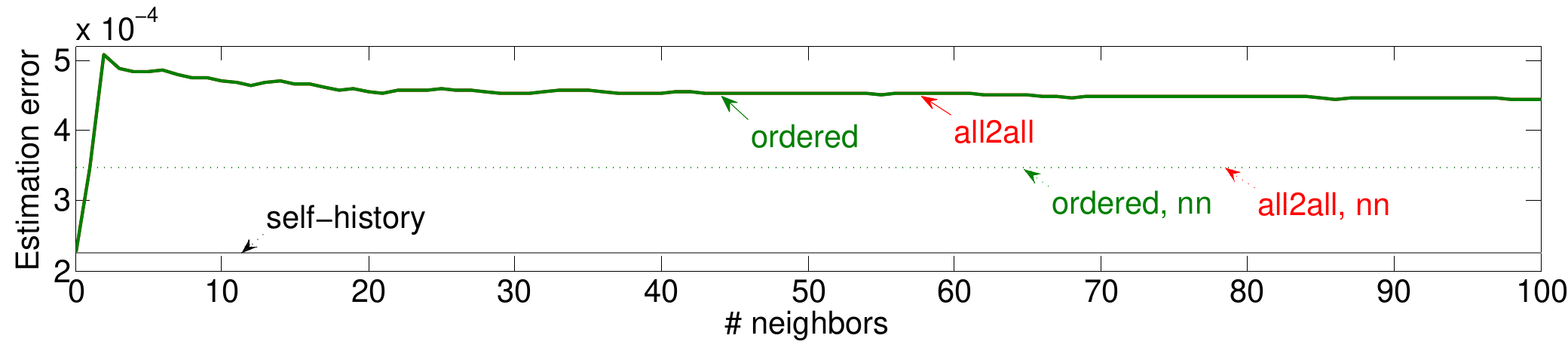}
    \label{fig:all2all_2}
  }
  \\
\hspace{-3mm}
  \subfigure[L=4]
  {
    \includegraphics[width=0.24\textwidth]{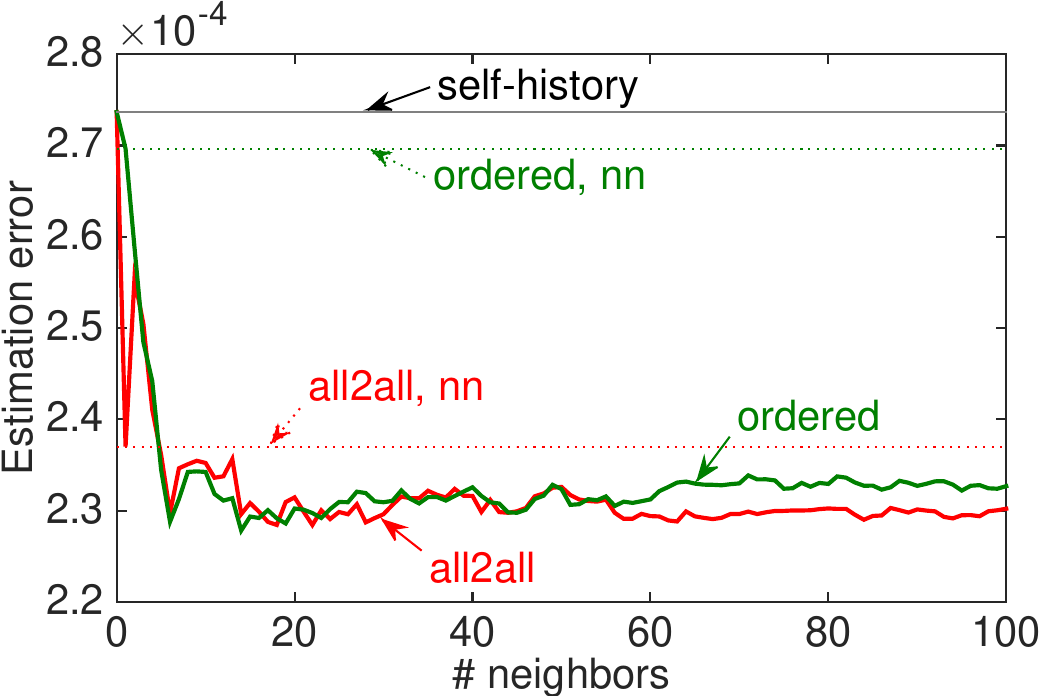}
    \label{fig:all2all_4}
  }
  \hspace{-5mm}
  \subfigure[L=6]
  {
    \includegraphics[width=0.24\textwidth]{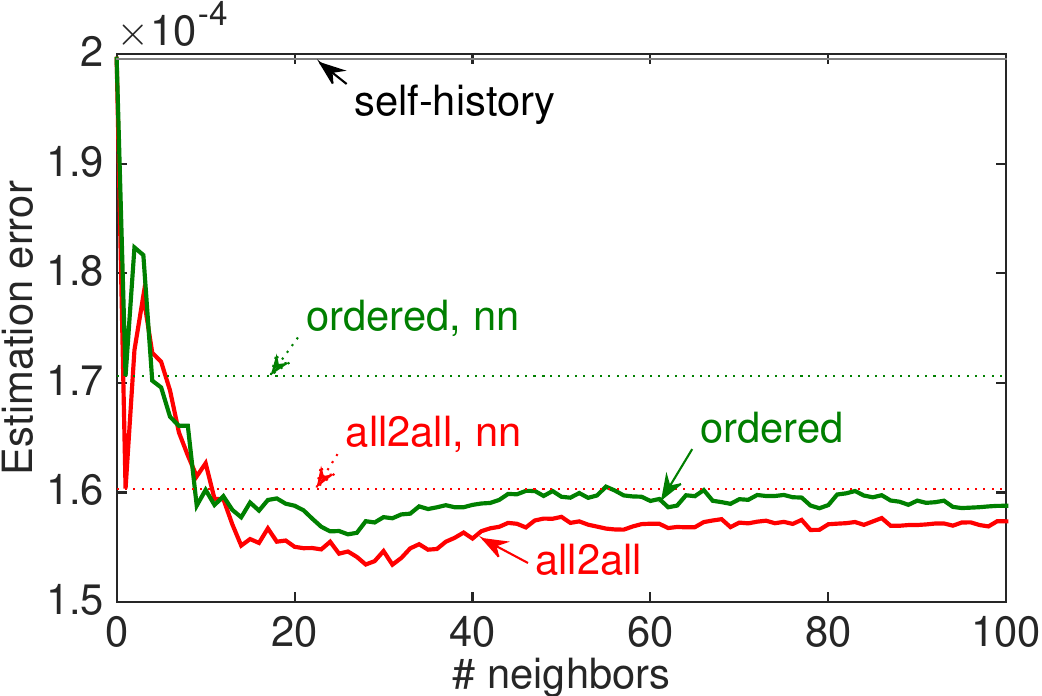}
    \label{fig:all2all_6}
  }
  \\
  \hspace{-3mm}
  \subfigure[L=8]
  {
    \includegraphics[width=0.24\textwidth]{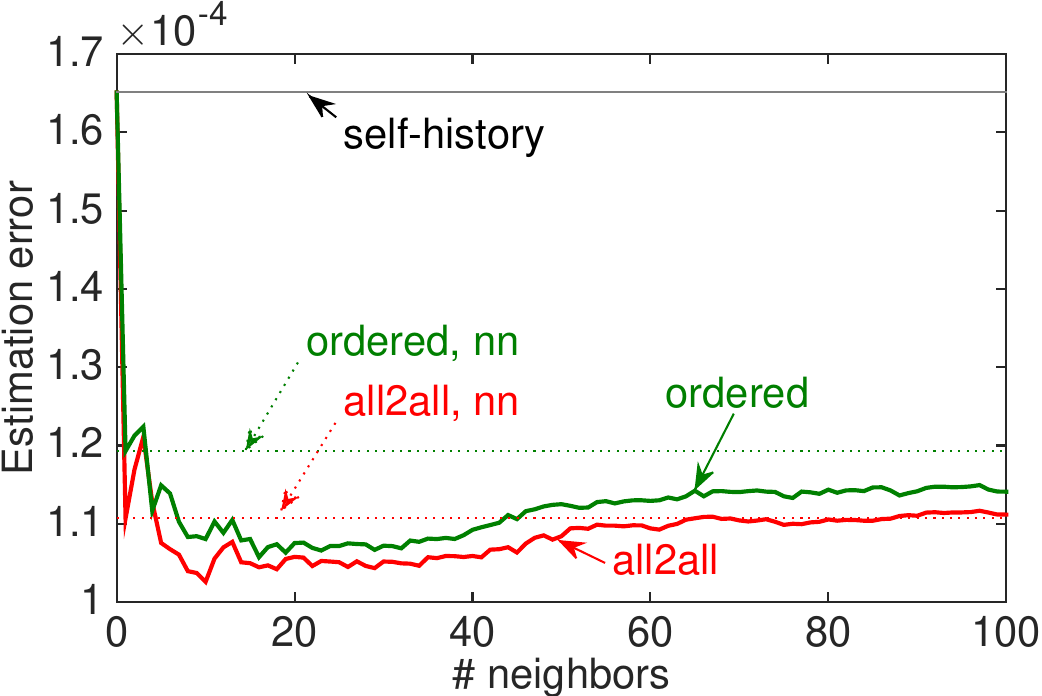}
    \label{fig:all2all_8}
  }
  \hspace{-5mm}
  \subfigure[L=10]
  {
    \includegraphics[width=0.24\textwidth]{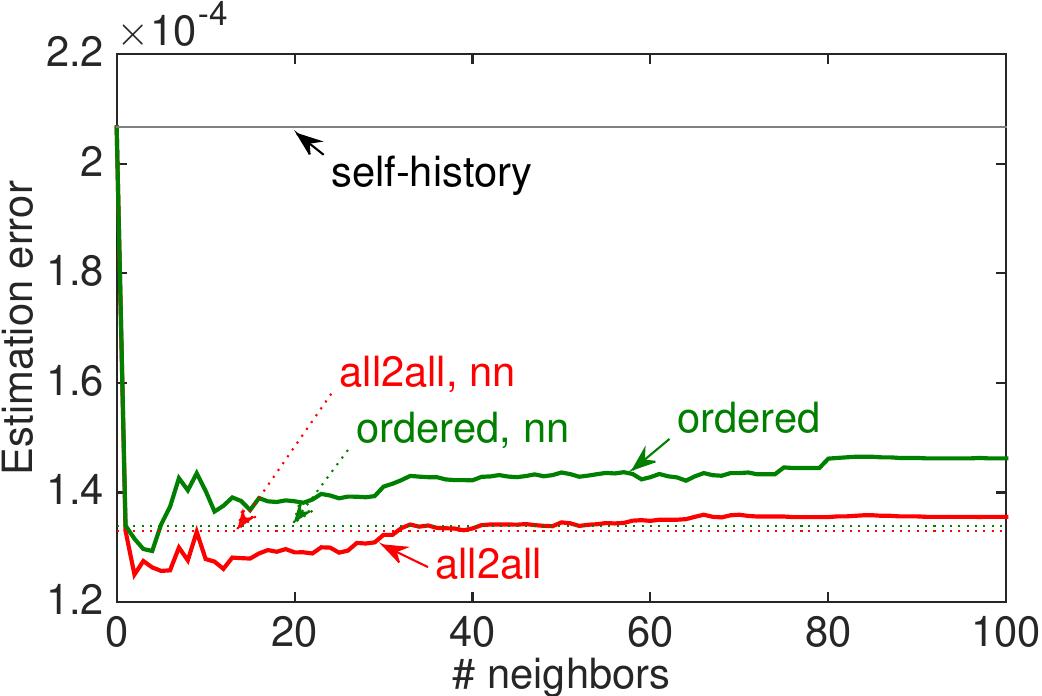}
    \label{fig:all2all_10}
  }
  \caption{Estimation error on entities with even $L$.
 }
  \label{fig:Leven}
\end{figure}

\begin{figure}[thb!]
  \centering
  \hspace{-3mm}
  \subfigure[all2all]
  {
    \includegraphics[width=0.24\textwidth]{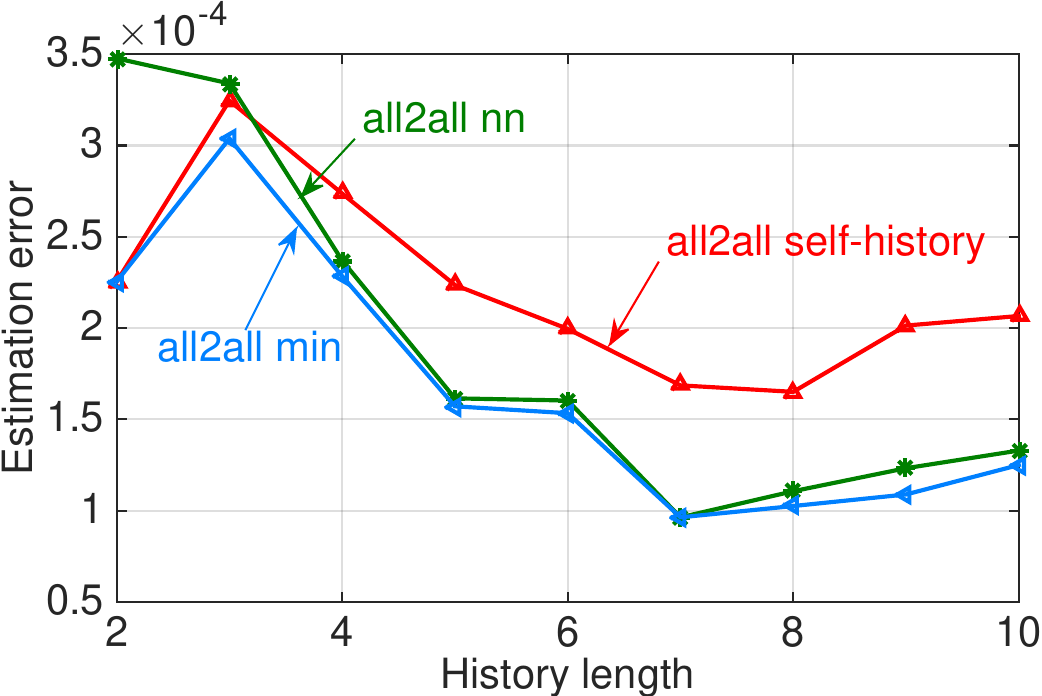}
    \label{fig:all2all_varlen}
  }
\hspace{-5mm}
  \subfigure[ordered]
  {
    \includegraphics[width=0.24\textwidth]{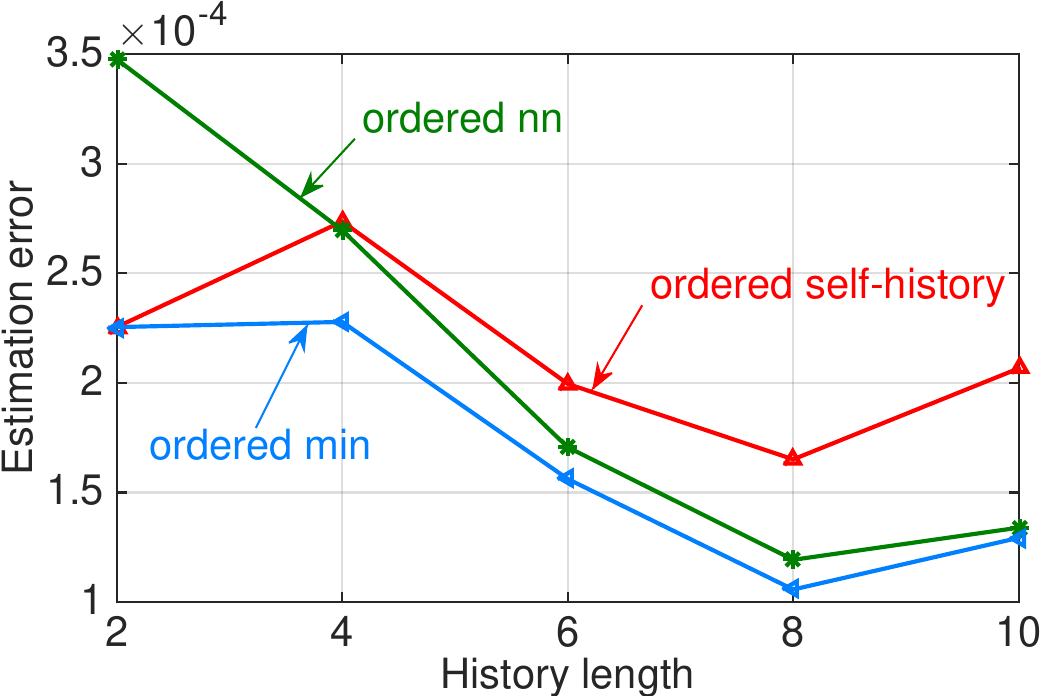}
    \label{fig:ordered_varlen}
  }
  \caption{Estimation error of different experimental settings.
  }
  \label{fig:Leven2}
\end{figure}

\begin{figure*}[tbh!]
  \centering
  \hspace{-3mm}
  \subfigure[L=3]
  {
    \includegraphics[width=0.24\textwidth]{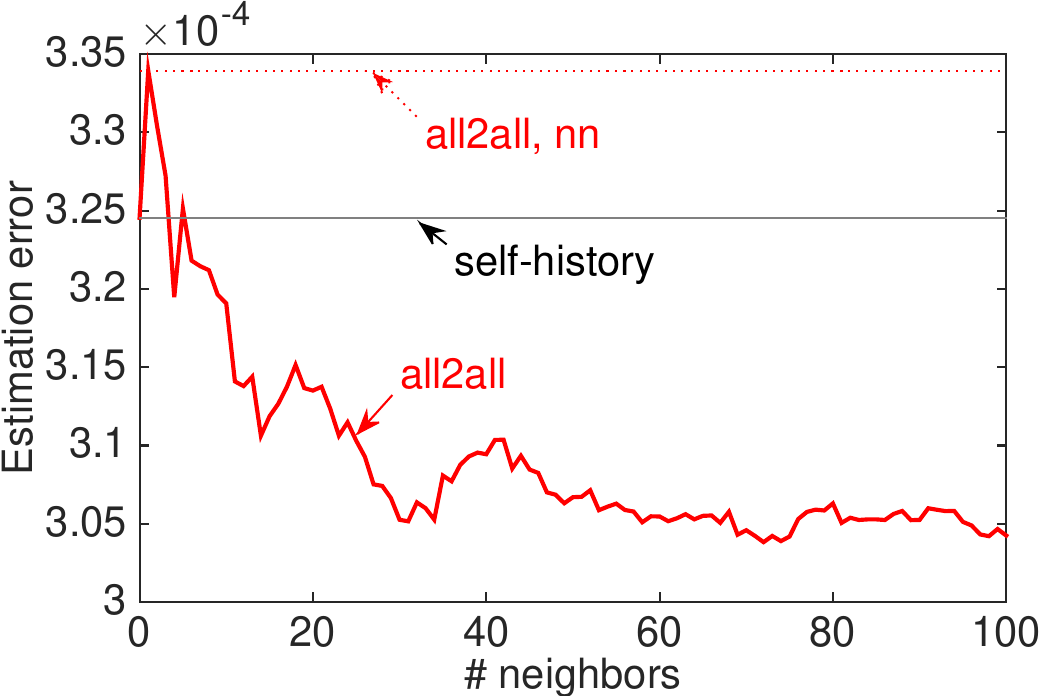}
    \label{fig:all2all_3}
  }
  \hspace{-3mm}
  \subfigure[L=5]
  {
    \includegraphics[width=0.24\textwidth]{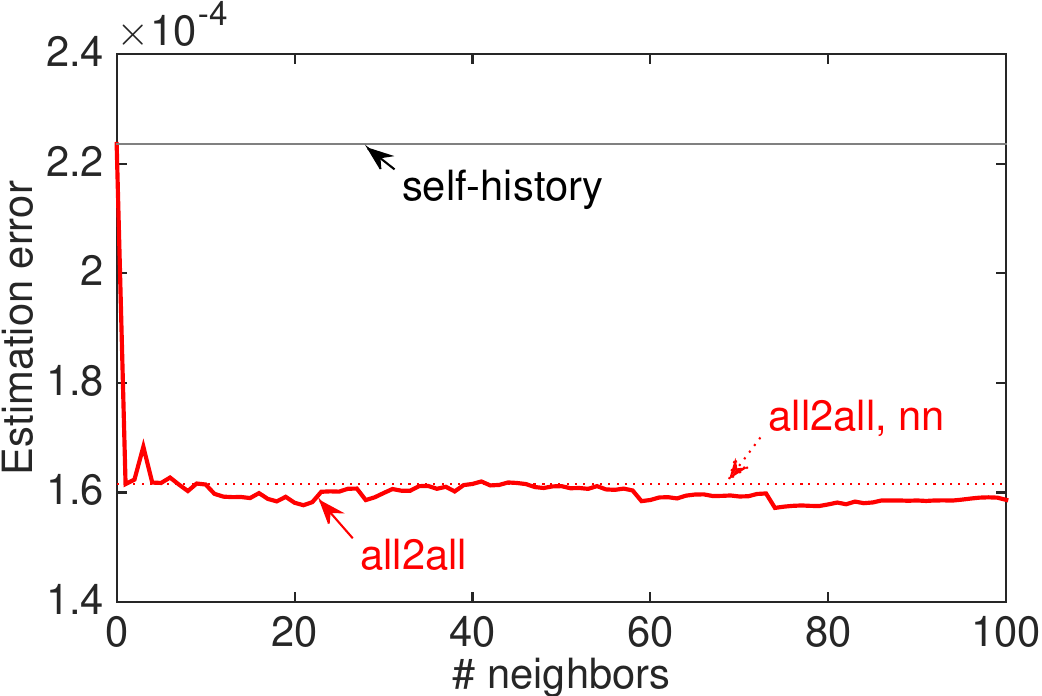}
    \label{fig:all2all_5}
  }
  \hspace{-3mm}
  \subfigure[L=7]
  {
    \includegraphics[width=0.24\textwidth]{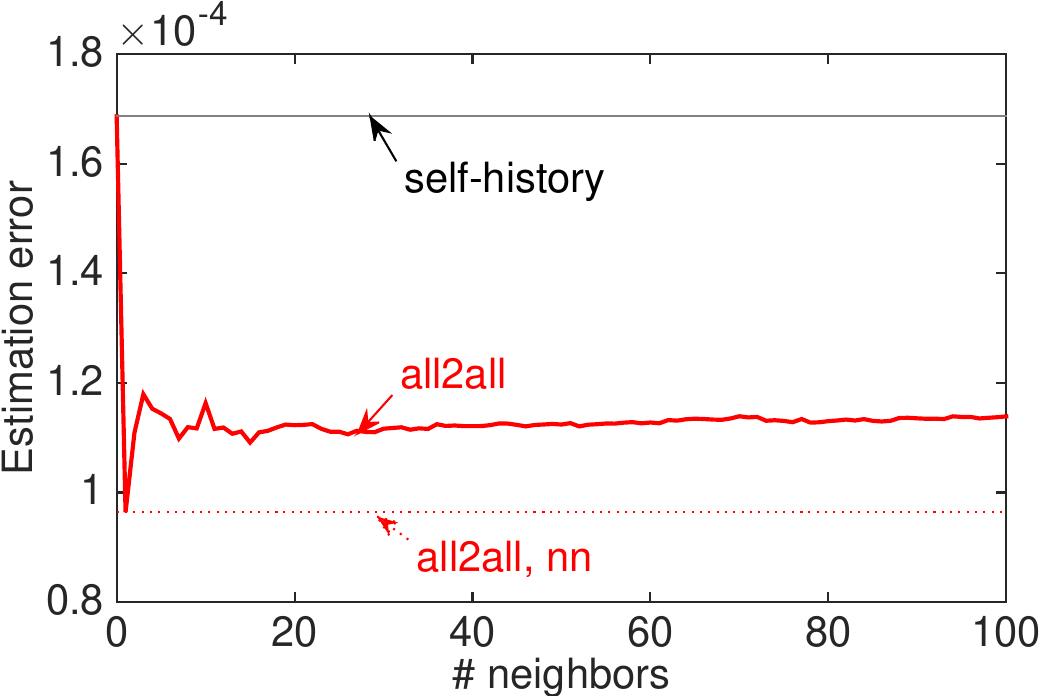}
    \label{fig:all2all_7}
  }
  \hspace{-3mm}
  \subfigure[L=9]
  {
    \includegraphics[width=0.24\textwidth]{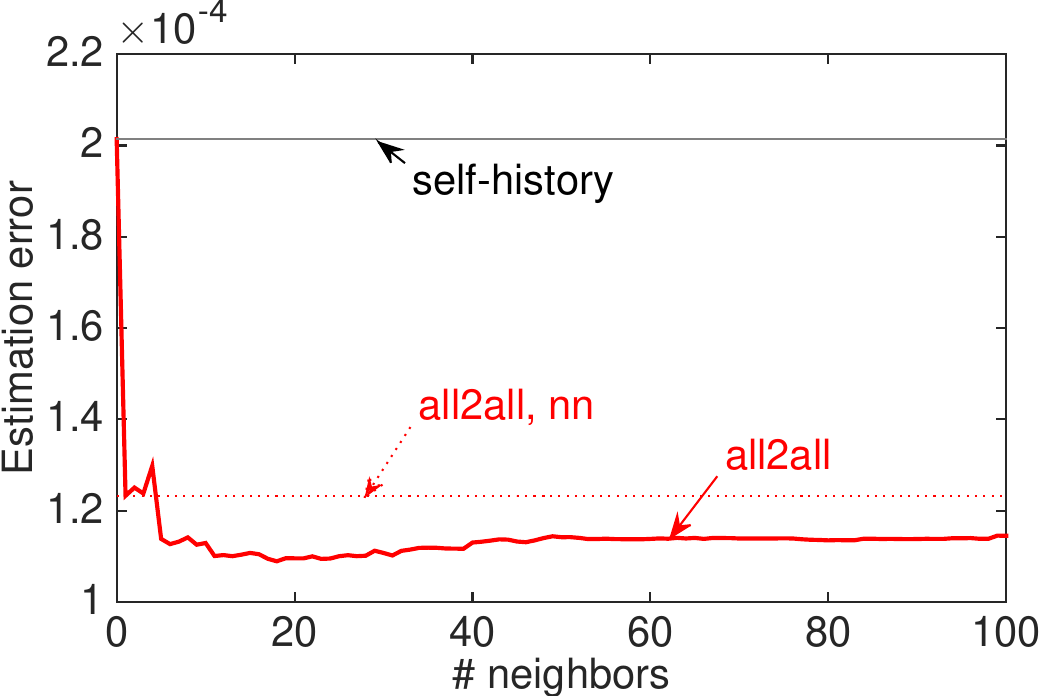}
    \label{fig:all2all_9}
  }
  \caption{Estimation error on entities with different lengths of trip history when $L$ is an odd number (i.e., $L=3, 5, 7, 9$).
  }
  \label{fig:Lodd}
\end{figure*}

\section{Experiments}

In this section, we conduct extensive empirical evaluation of our approach on a real-world public transportation management platform, 
and demonstrate its superior performance compared to the alternative baselines (i.e., prediction using ``singleton'' entity features, and prediction using the single nearest neighbor together with self history). In specific, we show that our framework can improve the performance of existing trip prediction algorithms via our similarity-based data refinement process.
Moreover, we investigate the impact of transforming the origin and destination features to a new set of features via non-negative matrix factorization.

\subsection{Experimental Setup}

 \paragraph{Dataset} We perform our experiments on the real-world trip specification data \texttt{Nancy2012} collected from the city of Nancy in France~\cite{ID/20120158,Chidlovskii:MUD22015}. This data is prepared from e-card validation collection. We query trip histories with different lengths, i.e. $L=2,3,4,5,6,7,8,9,10$, to produce different datasets.\footnote{Note that the  \emph{ordered} variant requires that the two trip histories (training and validation sets) must be aligned, i.e. they should have the same lengths. Thus, we perform $L=3,5,7,9$ only for \emph{all2all} variant.}
%
%
For each $L$ we collect $2,000$ entities from the database, unless there are less entities for a specific $L$ (For $L = 10$, we could collect only $740$ entities). We consider single-leg trips in our evaluations.\footnote{Note that over 95\% of the trips in our dataset are single leg.} Thus, each trip is specified by four elements: the longitude and the latitude of the origin and the longitude and the latitude of the destination.

We split each dataset into training and validation sets. Moreover, we have an additional trip (\emph{test} trip) for each entity which will be used as the ground truth (i.e. $T_{ut}^{tst}$) in order to investigate the correctness of our estimation/prediction.

\paragraph{Evaluation criteria} We compare the ground-truth and the predicted trips and compute the mean squared error 
\begin{equation}
  \hat{err} = \frac{1}{|\{\langle u,t\rangle\}|} \sum_{\langle u,t\rangle}\seuc(r_{ut} , T_{ut}^{tst}) \,,
\end{equation}
where $|\{\langle u,t\rangle\}|$ shows the number of test cases (entities).

\subsection{Numerical Analysis}
\paragraph{Results with different trip history lengths}
Figures~\ref{fig:Leven},~\ref{fig:Leven2} and~\ref{fig:Lodd}
illustrate the estimation error of computing the neighbors respectively for $L=2,4,6,8,10$ (i.e. when $L$ is an even number) and $L=3,5,7,9$ (i.e. when $L$ is an odd number). The neighbors are sorted according to their usefulness on validation set (i.e., their distance to the entity of interest, as defined in Equation~\eqref{eq:dist_ordered} and Equation~\eqref{eq:dist_all2all}). We investigate different number of neighbors per user, where \emph{no. of neighbors}$=0$, indicating the use of only self history for computing representative trip and prediction. Thus, this setting constitutes our baseline. Another baseline is to use the single nearest neighbor with self history. In Figure~\ref{fig:all2all_varlen} and Figure~\ref{fig:ordered_varlen}, we plot the prediction error using self-history only, nearest-neighbor, and the \emph{optimal set} of neighbors, respectively, for the two options of distance function. We observe,
i) except for $L=2$, our approach always leads to reducing the estimation error. For $L=2$, there is only one trip for training and one for validation. Thus, due to noise and sparsity, we are not able to select informative and reliable neighbors. However, once we increase the number of trips for training and validation sets, e.g. $L=3,4,5,6,7,8,9,10$, then, our approach yields computing better neighbors and  a better  representative trip among them, which thereby leads to reduce the estimation error by 15\%$\sim$40\%.
ii) As we increase the number of trips in history, i.e. the $L$, then we can better compute the neighbors and obtain a more reliable representative trip. Thus, a larger $L$ yields better performance in trip prediction, as well as smoother plots.
iii) The results are very much consistent between \emph{all2all} and \emph{ordered} variants, which  also indicates lack of any significant temporal  trip behavior. However, the advantage of the \emph{all2all} variant is that it can be employed even when there are entities with varying number of trips. Thus, the \emph{all2all} variant can replace whenever the \emph{ordered} variant is desired. Moreover,
\emph{all2all} in general outperforms \emph{ordered}, if we are allowed to include a fixed number of neighbors for each entity. Such observation suggest that we should consider to use \emph{all2all} whenever possible as the more reliable similarity measurement. We note that in the case of superiority of the \emph{ordered} variant, methods such as recurrent neural networks could be employed to better extract the temporal aspects of driving/transport trajectories \cite{DemetriouARC20,HoseiniRC21}.

\begin{figure}[htb]
  \centering
  \hspace{-3mm}
  \subfigure[L=6]
  {
    \includegraphics[width=0.24\textwidth]{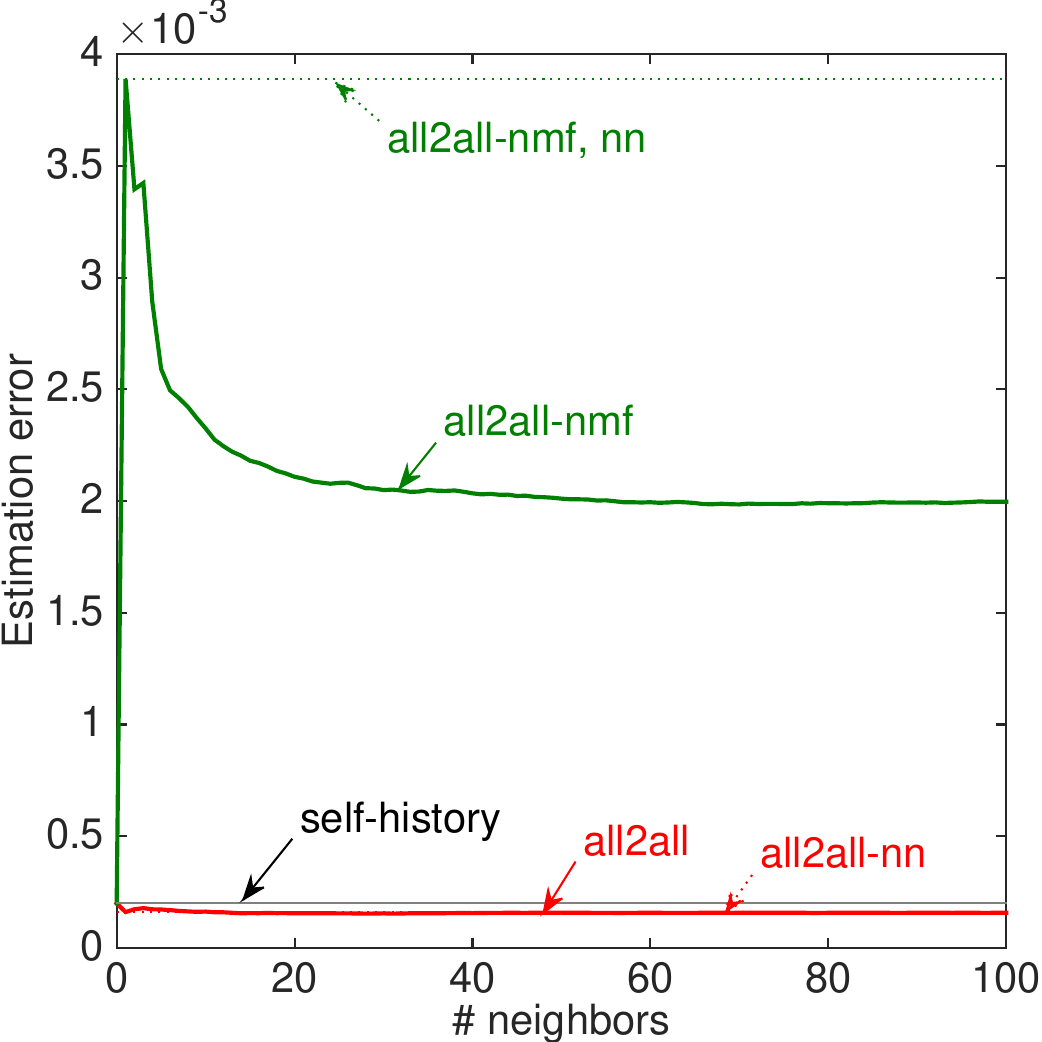}
    \label{fig:all2all_6_nmf}
  }
  \hspace{-5mm}
  \subfigure[L=8]
  {
    \includegraphics[width=0.24\textwidth]{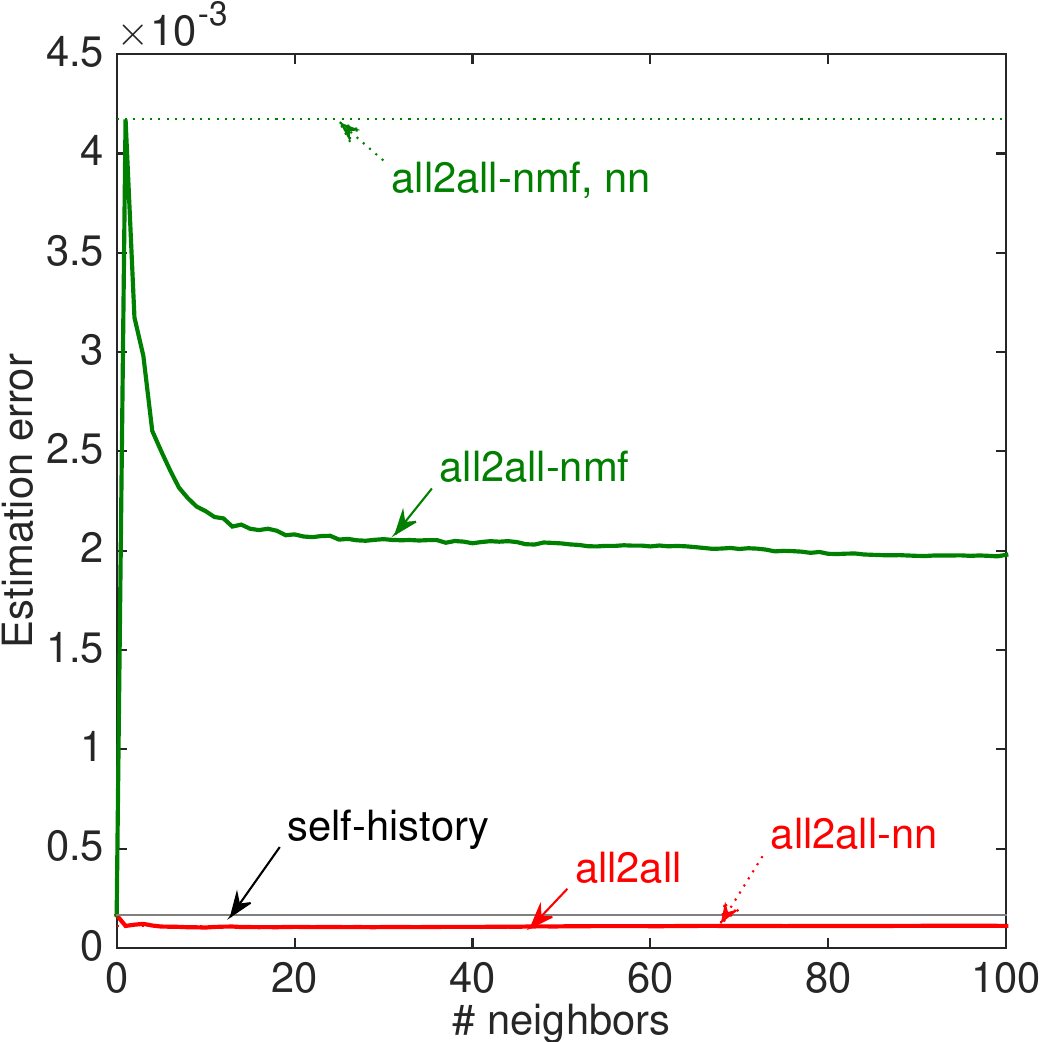}
    \label{fig:all2all_8_nmf}
  }
  \caption{Estimation error on entities with/without non-negative matrix factorization, for $L=6,8$.
}
  \label{fig:all2all_nmf}
\end{figure}

\paragraph{Pre-processing trips via non-negative matrix factorization} We then investigate that how the use of matrix factorization methods affects the prediction accuracy. In particular, we perform non-negative matrix factorization on the feature matrices, in order to transform the original features into another type of features which might be more suitable. This technique is very common in recommendations and collaborative filtering. We repeat the experiments for different number of hidden components and choose the best results. In our experiments, the optimal number of components is $4$. The results are shown in Figure~\ref{fig:all2all_nmf} for $L=6$ and $L=8$. For the other values of $L$, we observe very consistent results. We observe that transforming the original features into the new features leads to a significant increase in the prediction error. This observation implies that the original features are sufficient and informative enough to be used for the purpose of learning and prediction. Intuitively, this makes sense, because the original features are orthogonal (non-redundant) and sufficiently describe the origin and destination points.
%

\begin{figure}[th!]
  \centering
    \includegraphics[width=0.38\textwidth]{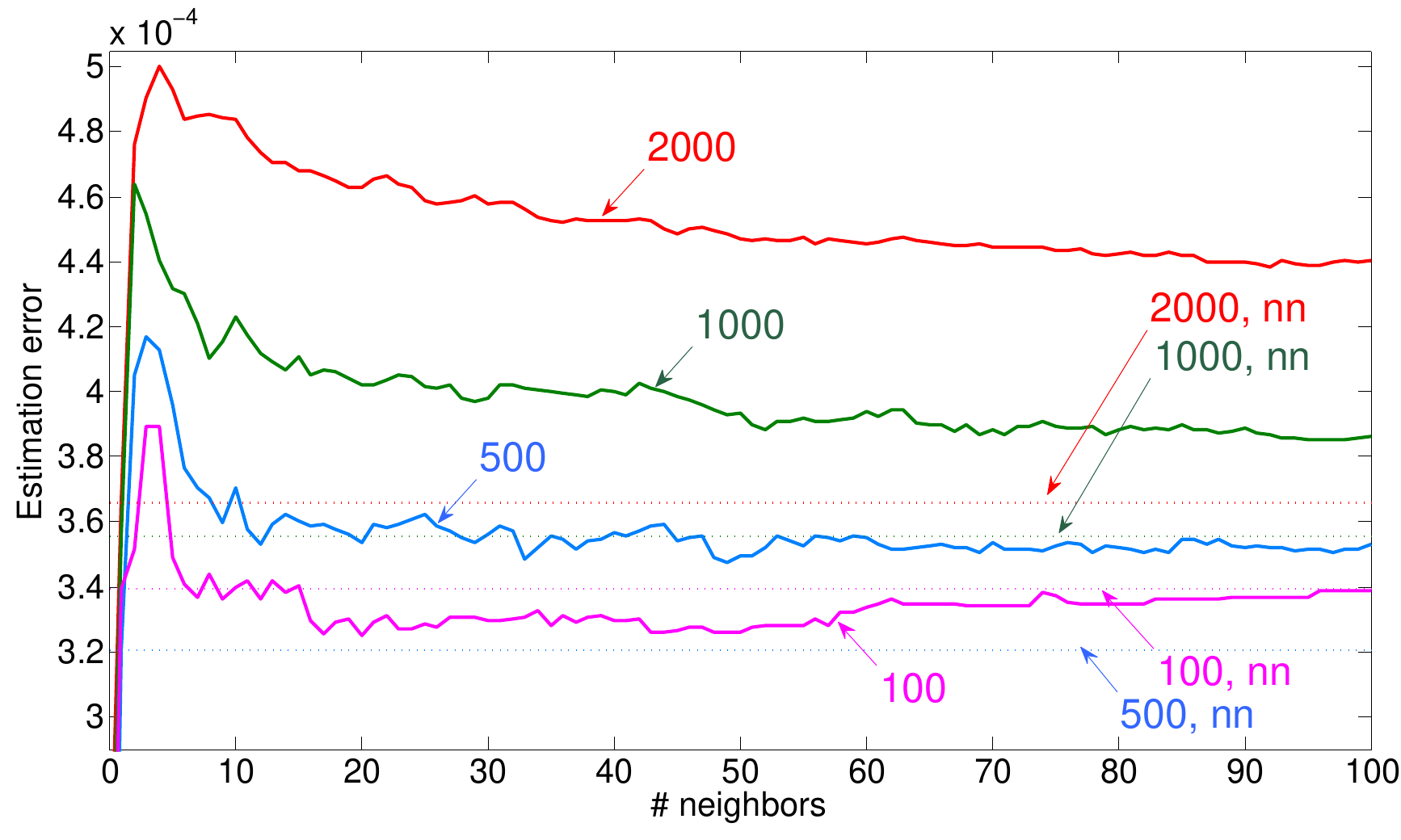}
  \caption{Estimation error when different number of entities with $L=2$ are augmented with $2000$ entities with $L=8$.
  }
  \label{fig:L2vsL8}
\end{figure}

\begin{figure}[ht!]
  \centering
  \hspace{-3mm}
  \subfigure[$L=3,4,5,6$]
  {
    \includegraphics[width=0.242\textwidth]{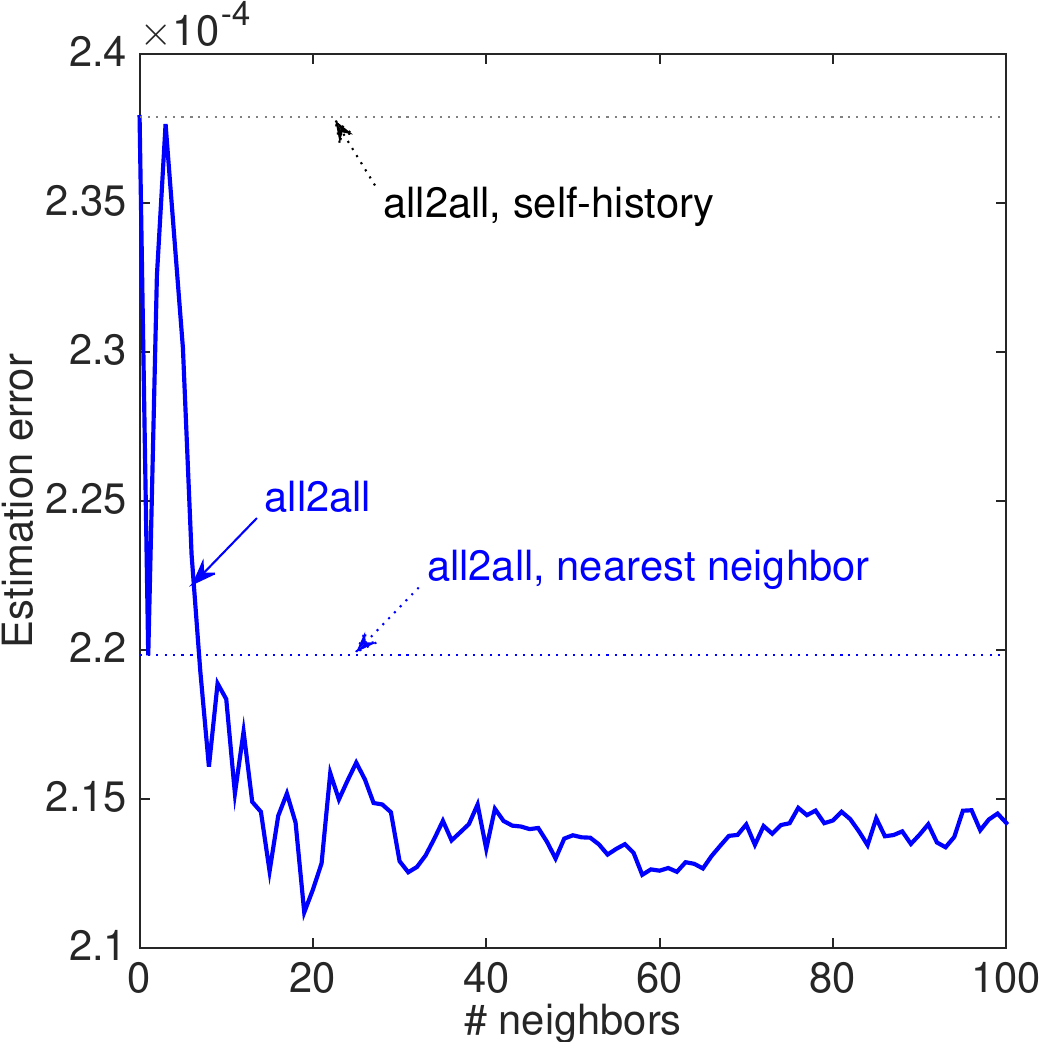}
    \label{fig:all2all_3_4_5_6_500}
  }
  \hspace{-5.4mm}
  \subfigure[$L=7,8,9,10$]
  {
    \includegraphics[width=0.242\textwidth]{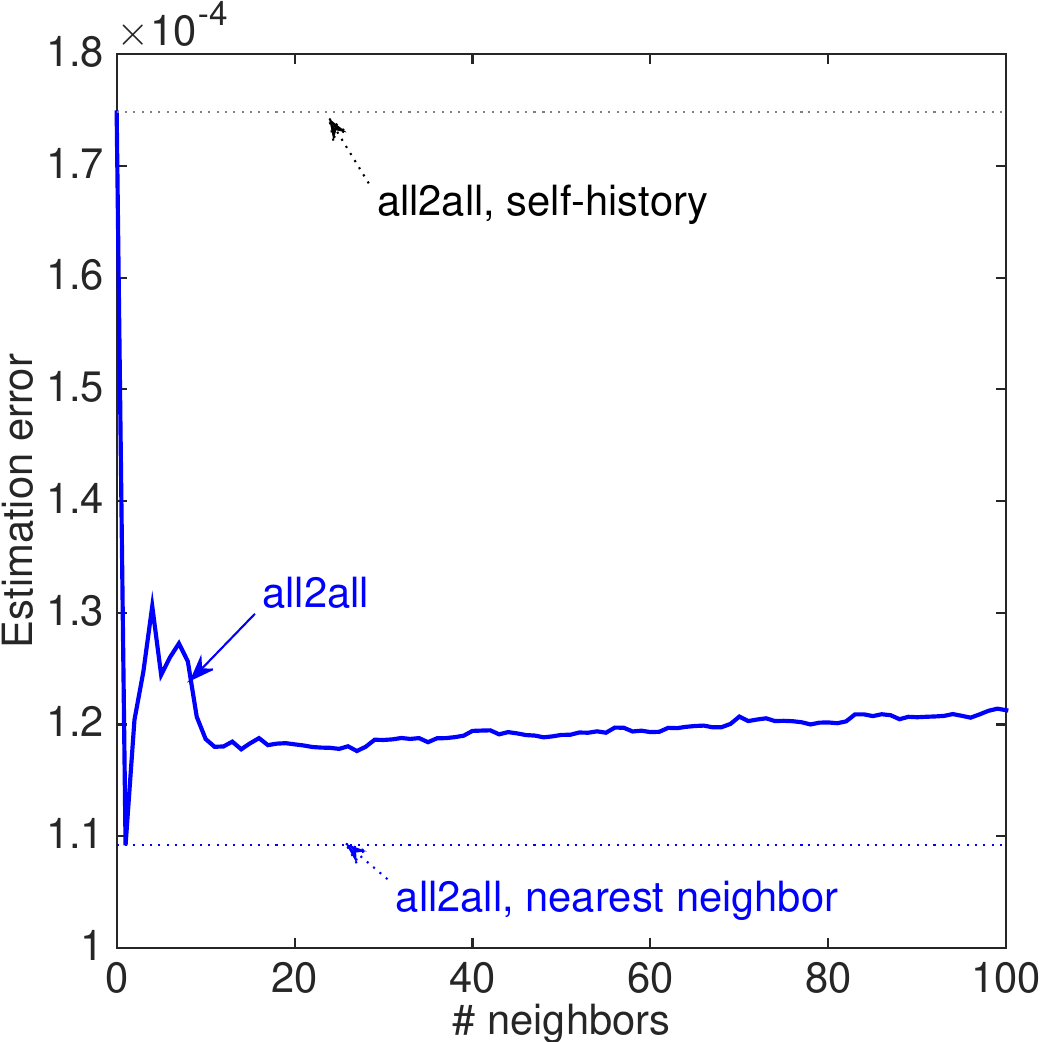}
    \label{fig:all2all_7_8_9_10_500}
  }
  \caption{Estimation error when entities with different number of trips are combined.
  }
  \label{fig:all2all_mixed}
\end{figure}

\paragraph{Enriching short-history entities  with  longer history neighbors} In the following, we investigate whether augmenting short histories with long histories can help to predict more accurate trips. In particular, we study the case of $L=2$, i.e. the only case where our approach fails to improve prediction accuracy. We choose different number of entities with $L=2$ (e.g. $100$, $500$, $1000$ and $2000$) and combine them with $2000$ entities whose $L$ is $8$. That is, we merge the training set (resp. validation set) of trips with $L=8$ with training set (resp. validation set) of trips with $L=2$. Figure~\ref{fig:L2vsL8} illustrates the results. We compute the estimation error only for the entities with $L=2$. We observe that,
i) the impact of very short histories (i.e. $L=2$) is very crucial, such that even augmenting them by very long histories dose not help much. We still see that using only the self-histories is a better choice for this particular case.
ii) As we increase the ratio of the number of long histories to the number of short histories, then we obtain better and more reliable neighbors such the estimation error decreases. We particularly observe this behavior when the number of entities changes from $2000$ to $1000$, $500$ and finally to $100$.

\paragraph{Combining entities with different number of trips} Finally, we consider combination of entities with different number of trips, i.e. with varying $L$. In the first case (Figure~\ref{fig:all2all_3_4_5_6_500}) we combine entities with $L=3,4,5,6$ trips, where the dataset contains $500$ entities from each category. In the second case (Figure~\ref{fig:all2all_7_8_9_10_500}), we consider $L=7,8,9,10$, and collect $500$ entities from each category.  For this setting, we employ \emph{all2all} measure for computing appropriate neighbors, since the entities have different number of trips. Figure~\ref{fig:all2all_mixed} illustrates the results.
We observe,
i) in both cases, our method helps to compute appropriate neighbors and thereby to reduce the estimation error.
ii) For the second case, i.e. when $L=7,8,9,10$, the estimation error is smaller (and smoother) than the first case. The reason is that in the second case the trip histories are longer, thus the representative trip can be computed in a more robust way.

\section{Discussion and Future Work}

\begin{figure}[!t]
  \centering
  \hspace{-3mm}
  \subfigure[Public transportation routes]
  {
    \includegraphics[width=0.24\textwidth]{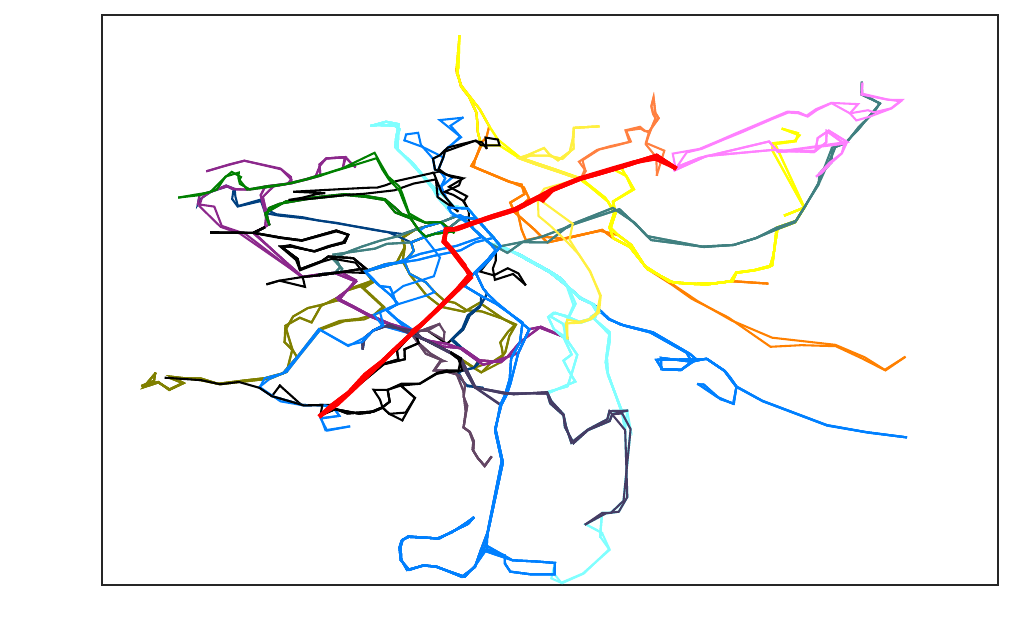}
    \label{fig:allstops}
  }
   \hspace{-5mm}
  \subfigure[Example trips]
  {
    \includegraphics[width=0.24\textwidth]{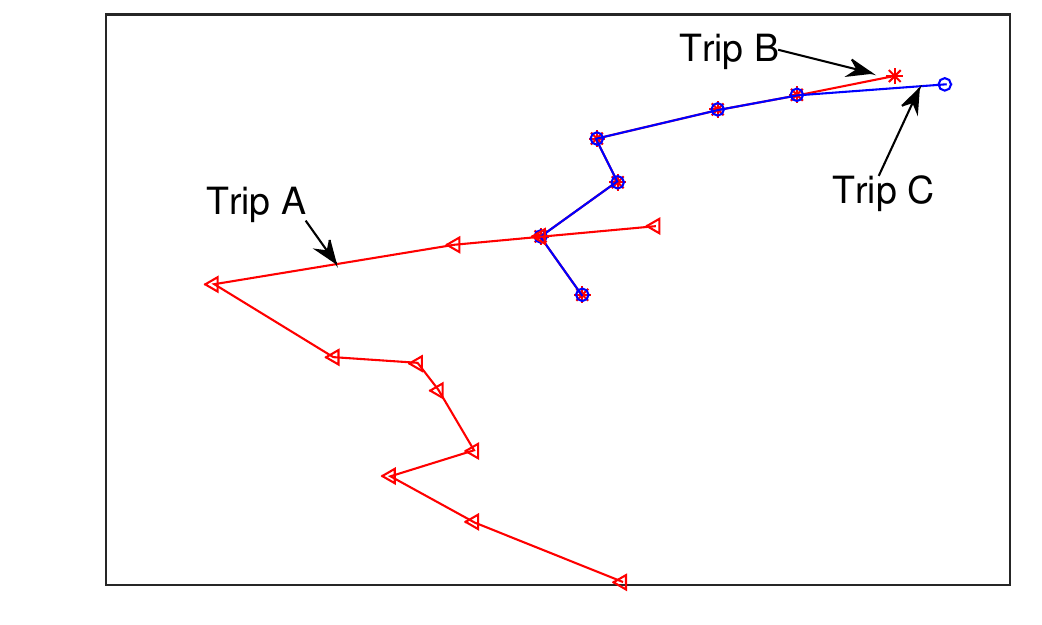}
    \label{fig:examplestops}
  }
  \caption{Routes and Stops in \texttt{Nancy2012}.
  }
  \label{fig:stops}
\end{figure}

In this section, we discuss some potential directions on improving the prediction accuracy. One defining factor for performance is the trips' initial feature representations. As previously discussed, the performance of the predictor relies on our definition of the distance function $dist(.,.)$, which currently is defined as a function of the (pairwise-) squared Euclidean distances between trips. However, the geographical information about a trip is more than just the origin and destination stop. As demonstrated in Figure~\ref{fig:stops}, straight line distance between $(o,d)$ pairs hardly reflects the scales of the difference between different trips. In Figure~\ref{fig:examplestops}, Trip $B$ and Trip $C$ represent the same service line in different hours of the day. They are almost identical except for the last stop. Trip $A$ and Trip $B$ (or $C$) are very different, though they still share a common stop which could be a popular transit stop for 2-leg trips (i.e., some users travel on Trip $A$ may transfer to $B$ (or $C$) at the intersecting point). To capture such potentially useful information, we propose a new distance measure between trips, $\tripd(.,.)$, defined as follows:
\begin{align}
  \tripd(p_i, q_i) = \left(1-\frac{p_i\cap q_i}{p_i\cup q_i}\right) * \seuc(p_i, q_i)
\end{align}
where the first term on the R.H.S. represents the Jaccard distance between trip $p_i$ and $q_i$ if we view them as sets of intermediate stops. This heuristic captures the intuition that if two trips share many common stops, even though the ending stops are far apart, we may still want to treat them as ``somewhat similar'' since they can belong to different segments of the same service line, or the two trips can be potential transfer trips for each other.

Since our method for proposing neighbors is orthogonal to the feature extraction component, we may preprocess the current features by transferring them into more robust, noise-resilient features, via commonly used techniques such as non-negative matrix factorization or truncated SVD. This black-box feature engineering component might be particularly helpful when we use more complicated features or a combination of different criteria. Another choice would be  minimax distance which computes the transitive relations and elongated patterns in a nonparametric way  \cite{ChehreghaniMLJ20,ChehreghaniECIR17}.

\section{Conclusion}

We propose a new method for trip prediction by taking into account the user trip history. We augment users' trip history with trips taken by similar users, where the similarity between users are directly guided by the prediction error. We show that by incorporating the augmented trip history, one can improve the informativeness of users' self histories, and hence improve the overall performance of the trip prediction system. 
We perform experiments on a real-world dataset collected from real trip transactions of the passengers in the city of Nancy in France.

\paragraph*{Acknowledgment}
The work was mainly done at Xerox Research Centre Europe.
We thank Boris Chidlovskii
 for the insightful discussions and his  help in  working with the \texttt{Nancy2012} dataset.

\bibliographystyle{plain}
\bibliography{references}

\end{document}